\documentclass[conference]{IEEEtran}
\IEEEoverridecommandlockouts
\usepackage{cite}
\usepackage{amsmath,amssymb,amsfonts}
\usepackage{algorithmic}
\usepackage{graphicx}
\usepackage{textcomp}
\usepackage{xcolor}
\def\BibTeX{{\rm B\kern-.05em{\sc i\kern-.025em b}\kern-.08em
		T\kern-.1667em\lower.7ex\hbox{E}\kern-.125emX}}
\begin{document}

\title{LLM4XCE: Large Language Models for Extremely Large-Scale Massive MIMO Channel Estimation
	\thanks{This work was supported in part by the National Science Foundation of
		China under Grants 62471226, 62101253, the Fundamental Research Funds for the Central Universities (NO. NS2024023), and the Natural Science
		Foundation of Jiangsu Province under Grants BK20210283. (Corresponding author: Peihao Dong.)}
}

\author{\IEEEauthorblockN{Renbin Li, Shuangshuang Li, Peihao Dong}
	\IEEEauthorblockA{ College of Electronic and Information Engineering, Nanjing University of Aeronautics and Astronautics, Nanjing 211106, China\\
		Email: \{renbinli, lishsh, phdong\}@nuaa.edu.cn}
}

\maketitle

\begin{abstract}
Extremely large-scale massive multiple-input multiple-output (XL-MIMO) is a key enabler for sixth-generation (6G) networks, offering massive spatial degrees of freedom. Despite these advantages, the coexistence of near-field and far-field effects in hybrid-field channels presents significant challenges for accurate estimation, where traditional methods often struggle to generalize effectively. In recent years, large language models (LLMs) have achieved impressive performance on downstream tasks via fine-tuning, aligning with the semantic communication shift toward task-oriented understanding over bit-level accuracy.
 Motivated by this, we propose Large Language Models for XL-MIMO Channel Estimation (LLM4XCE), a novel channel estimation framework that leverages the semantic modeling capabilities of large language models to recover essential spatial-channel representations for downstream tasks. The model integrates a carefully designed embedding module with Parallel Feature-Spatial Attention, enabling deep fusion of pilot features and spatial structures to construct a semantically rich representation for LLM input. By fine-tuning only the top two Transformer layers, our method effectively captures latent dependencies in the pilot data while ensuring high training efficiency. Extensive simulations demonstrate that LLM4XCE significantly outperforms existing state-of-the-art methods under hybrid-field conditions, achieving superior estimation accuracy and generalization performance. 
\end{abstract}

\section{Introduction}
With the advancement of sixth-generation mobile communication (6G), extremely large-scale massive multiple-input multiple-output (XL-MIMO) systems have emerged as a key enabling technology for high-capacity and highly reliable communications, owing to their massive antenna arrays, ultra-high spectral efficiency, and spatial resolution\cite{b1}, \cite{b2}. However, the accurate acquisition of channel state information (CSI) remains a major performance bottleneck, especially in hybrid-field environments involving near-field and far-field coexistence, rich multipath propagation and fine-grained beam control. In such cases, traditional channel estimation methods often fail to deliver satisfactory performance. 

In recent years, deep learning-based methods have been introduced to channel estimation tasks\cite{b3}. In\cite{b4}, a lightweight deep convolutional network was designed specifically for XL-MIMO systems to improve estimation efficiency while reducing computational overhead. In\cite{b5}, an attention-aided deep learning framework was proposed by integrating attention mechanisms into conventional deep networks, which significantly improved estimation accuracy. Furthermore, in\cite{b6}, a hybrid attention-enhanced Transformer was developed to model the complex channel structure in XL-MIMO, effectively integrating multi-dimensional feature attention. However, these methods still suffer from degraded estimation accuracy in complex scenarios, limited adaptability to environmental variations, and poor generalization capability.

Meanwhile, large language models (LLMs)~\cite{b7} have started to be applied to wireless communication tasks, such as time series forecasting, channel prediction, and beam prediction, showing great potential in data-driven modeling and cross-task generalization. For example, in~\cite{b8}, the Time-LLM method reprogrammed input sequences to apply pre-trained language models to time series forecasting, achieving promising results on benchmark datasets. In~\cite{b9}, the channel prediction framework discretized CSI sequences into communication language and modeled them using LLMs to capture context-aware temporal dependencies. In~\cite{b10}, beamforming indices were encoded as text tokens and classified using an LLM, leading to improved beam selection performance in millimeter-wave systems. On the other hand, semantic communication shifts the focus from bit-level accuracy to conveying task-relevant meaning~\cite{b11}. Semantic features have been demonstrated to be robust against channel impairments and beneficial for downstream tasks, as evidenced by recent studies \cite{b12}, \cite{b13}. 

Motivated by these advances, this paper proposes a novel LLM-based channel estimation framework that leverages semantic representations to improve robustness and generalization under hybrid-field channel conditions in XL-MIMO systems. The semantic representations in the proposed framework include target-related information such as angles and distances, as well as spatial structural features of the channel matrix, which help enhance the task relevance and interference robustness of channel modeling. The main contributions of this paper can be summarized as follows

1) This work is the first to adapt LLMs for hybrid-field channel estimation, and proposes a Parallel Feature-Spatial Attention module to effectively fuse the feature and spatial dimensions of pilot signals. 
This addresses the input mismatch between pilot matrices and LLMs. It also establishes a general input mapping framework that promotes efficient fusion and processing of multimodal data in semantic communication.

2) To significantly reduce training costs, a lightweight LLM fine-tuning strategy is designed that updates only the top two Transformer layers, effectively lowering the computational resources required for training. Simulation results show that the proposed method outperforms various advanced baselines in both estimation accuracy and generalization under complex hybrid-field scenarios.\textit{}\hspace{2em}

\section{System Model}

\subsection{Signal Transmission Model}
As shown in Fig. 1, we consider a time-division duplexing (TDD) XL-MIMO system, in which a base station (BS) equipped with 
$M$ antennas communicates with a single-antenna mobile user. The wireless channel between the BS and the user consists of 
$L$ propagation paths, capturing the multipath characteristics of the environment.

 The pilot signal transmitted by the user within a time slot is denoted by $ x \in \mathbb{C}$, the uplink received signals at the BS $\mathbf y$ can be formulated by
 	\begin{equation}
 	\mathbf{y} = \mathbf{h} x +\mathbf{n}\label{eq},
 \end{equation}
where $ \mathbf{h} \in \mathbb{C}^{M \times 1}$ as the channel vector between the base station and the user, and $ \mathbf{n} \sim \mathcal{C} \mathcal{N} \left( {0}, {1} \right) \in \mathbb{C}^{M\times1}
$ represents the additive Gaussian noise. 
\begin{figure}[htbp]
	\centering
	\includegraphics[width=0.44\textwidth]{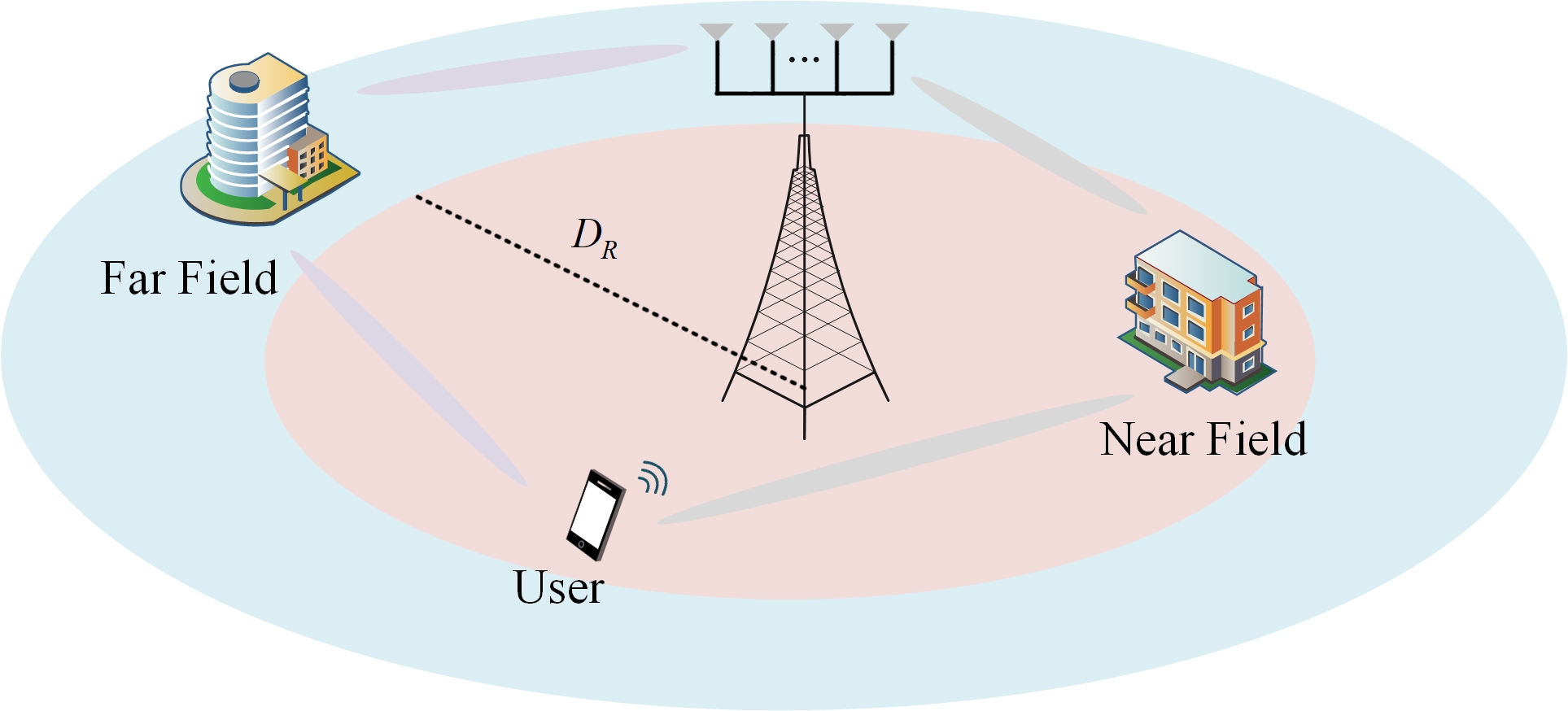}
	\caption{A hybrid-field XL-MIMO system.}
	\label{fig}
\end{figure}
\subsection{Channel Model}
In XL-MIMO systems, far-field propagation can typically be approximated as planar waves. However, in the near-field region, wave propagation exhibits spherical characteristics and thus cannot be simplified using the far-field assumption. Therefore, near-field effects must be explicitly considered. According to\cite{b14}, the boundary between the near-field and far-field regions is generally defined by the Rayleigh distance $D_{\mathrm{R}} = \frac{2 D_\mathrm{a}^2}{\lambda}$, where ${ D_\mathrm{a}}$ and $ {\lambda}$ represent the array aperture and the carrier wavelength, respectively. When the propagation distance between the array and the scatterer exceeds  $D_{{\mathrm{R}}}$
, the wavefront can be approximated as planar and the signal is considered to be in the far-field region. Otherwise, the wavefront is spherical and the signal lies in the near-field region. Assuming a uniform linear array (ULA) equipped with $M$ antennas, the spacing between adjacent antennas is set to $d=\frac{\lambda}2$, the Rayleigh distance is given by
\begin{equation}
	D_{\mathrm{R}} = \frac{2\left(M \frac{\lambda}{2} \right)^2}{\lambda} = \frac{M^2 \lambda}{{2} } .
\end{equation}

For instance, when an XL-MIMO BS is equipped with 256 antennas and operates at the frequency band of 30 GHz, $D_\mathrm{R}$ can reach 327 meters.
 Such a considerable range may cover a portion of users, placing them in the near-field region and consequently subjecting them to near-field propagation effects.

 When the scattering distance is larger than \( D_{\mathrm{R}} \), the channel is modeled using 
 the far-field formula expressed as
 \begin{equation}
 	\mathbf{h}_{\mathrm{f}} = \sqrt{\frac{M}{L}}  \sum_{l=1}^{L} g_{l} \mathbf{a}_{\mathrm{f}}(\theta_l), \label{eq3}
 \end{equation}
 where,  $\mathbf{a}_{\mathrm{f}}(\theta_l)\in \mathbb{C}^{M \times 1}$
 denotes the array response vector of the base station for the 
 $l$-th path, given by
  \begin{equation}
  	\scalebox{1}{$
 \mathbf{a}_{\mathrm{f}}(\theta_l) = \frac{1}{\sqrt{M}}[1, e^{-j 2\pi \frac{d}{\lambda} \sin \theta_l}, \ldots, e^{-j 2\pi \frac{d}{\lambda} (M-1) \sin \theta_l}]^T,$}
 \end{equation}
 where \( \theta_l \in [-\frac{\pi}{2}, \frac{\pi}{2}] \) and \(g_{l}\) denote the azimuth
 angle and the gain of the $l$-th path, respectively.
 
When the scattering distance is smaller than \( D_{\mathrm{R}} \), the channel is modeled using 
the near-field formula expressed as
 \begin{equation}
 	\mathbf{h}_{\mathrm{n}} = \sqrt{\frac{M}{L}} \sum_{l=1}^{L} g_l \mathbf{a}_{\mathrm{n}}(\theta_l, r_l)\label{eq},
 \end{equation}
 where  $ \mathbf{a}_{\mathrm{n}}(\theta_l, r_l) = \frac{1}{\sqrt{M}} [ e^{-j \frac{2\pi}{\lambda} (r_{l,1} - r_l)}, \ldots, e^{-j \frac{2\pi}{\lambda} (r_{l,M} - r_l)} ]^T
 $ is the antenna array response vector with the azimuth angle $\theta_l \in \left[ -\frac{\pi}{2}, \frac{\pi}{2} \right]
 $ and the distance $r_l $ for the $l$-th path. The $ g_l$ is the path gain and the distance $r_l $
 represents the distance from the $l$-th scatterer to the center of the base station antenna array. The distance between the $l$-th scatterer and the $m$-th antenna element is given by 
 $ r_{l,m} = \sqrt{r_l^2 + \delta_m^2 d^2 - 2 r_l \delta_m d \sin \theta_l}$, where $\delta_m = \frac{2m - M - 1}{2}, m = 1, \ldots, M
 $.
 
However, in certain practical scenarios, the scattering environment may include both near-field and far-field components. In such cases, the signal propagation and reception are simultaneously influenced by spherical wavefronts generated by near-field scatterers and planar wavefronts arising from far-field scatterers. This results in a hybrid-field channel model that deviates from traditional assumptions and introduces new challenges for both system design and signal processing. The hybrid-field channel model can be expressed as
\begin{equation}
	\begin{split}
		\mathbf{h }&= \mathbf{h}_{\mathrm{f}} + \mathbf{h}_{\mathrm{n}} \\
		&= \sqrt{\frac{M}{L}} \left[ \sum_{l=1}^{L_0} g_{l} \mathbf{a}_{\mathrm{f}}(\theta_l) + \sum_{l=L_0+1}^{L} g_{l} 
		\mathbf{a}_{\mathrm{n}}(\theta_l, r_l) \right].\label{equ5}
	\end{split}
\end{equation}
The above equation shows that the hybrid-field contains \( L_0 \) far-field paths and \( L - L_0 \) near-field paths.

	\begin{figure*}[t]
	\centering
	\includegraphics[width=1\textwidth]{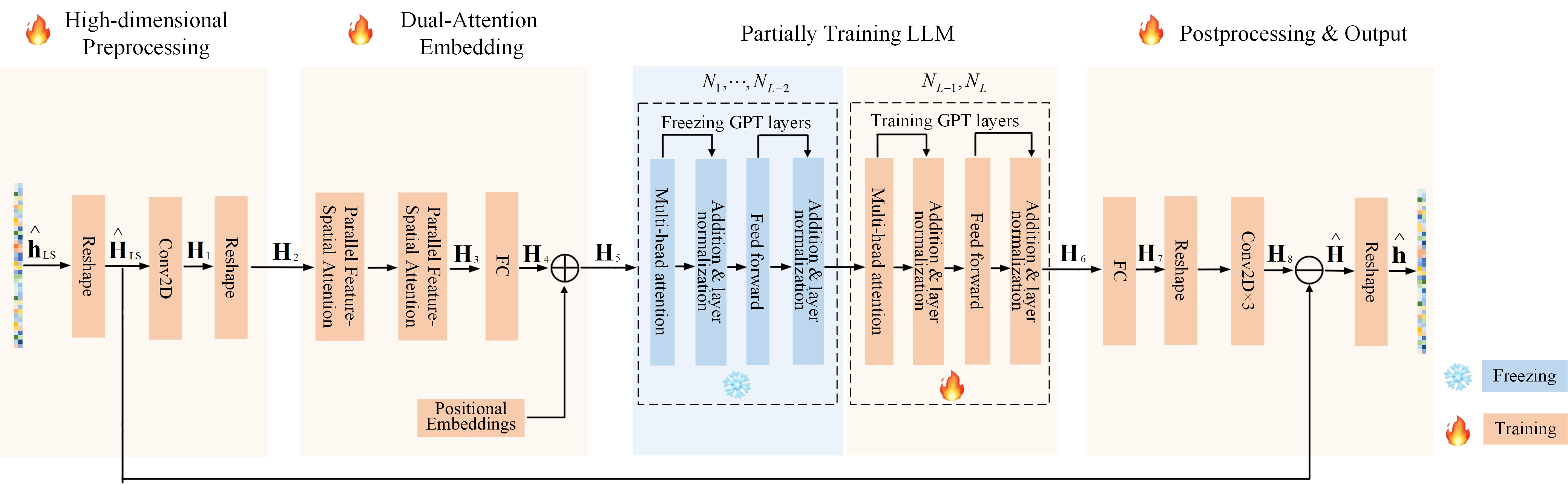}  
	\caption{Overview of the proposed LLM4XCE network architecture.}
	\label{figure2}
\end{figure*}
\section{LLM For Channel Estimation}
In this section, we propose a Large Language Model-based channel estimation method for extremely large-scale MIMO systems, named LLM4XCE, which aims to recover the true high-dimensional channel matrix from its noisy observation. To enable the text-pretrained LLM to effectively handle the complex-valued matrix format of the noisy channel, we design a dedicated format conversion and feature extraction framework, which includes High-dimensional Preprocessing module, Dual-Attention Embedding module, Partially Training LLM module, Residual structure, and Postprocessing \& Output module.
The architecture of the proposed network components is illustrated in Fig.~\ref{figure2}, and the following sections provide a detailed description of each network module.
\subsection{High-dimensional Preprocessing}\label{AA}
Since the channels of multiple users exhibit the same general structure as in equation (1), our design is readily extendable to the multi-user scenario. For simplicity, we concentrate on the single-user case in the subsequent analysis. Given that \( x \) in (1) is a pilot signal with a known power, and assuming its power is \( P \), we can express 
\( x\) as \( x = \sqrt{P} \bar{x} \), where \( \bar{x} \) is a pilot signal with unit power. Thus, without loss of generality, 
we can simplify (1) to
\begin{equation}
	\mathbf{y} = \sqrt{P}\mathbf{h} +\mathbf{n},
\end{equation}
based on this, the least squares (LS) estimate of  $\mathbf{ {h}} \in \mathbb{C}^{ {M\times 1}} $ is given by
\begin{equation}
	\hat{\mathbf{h}}_{\mathrm{LS}} = \frac{\mathbf{y}}{\sqrt{P}} = \mathbf{h} + \frac{\mathbf{n}}{\sqrt{P}}.
\end{equation}

The coarse estimate $\hat{\mathbf{h}}_{\mathrm{LS}}$ will be refined by LLM4XCE to yield a more accurate version. The complex vector ${\mathbf{h}} $ is split into its real and imaginary components of equal dimensions, where are then concatenated and reshaped into a tensor with dimensions $\mathbf{ {H}} \in \mathbb{R}^{\sqrt{M} \times \sqrt{M}\times  2} $, where $M$ is assumed to be a perfect square.
Similarly, the complex vector $	\hat{\mathbf{h}}_{\mathrm{LS}} $ is reshaped into a matrix $\hat{\mathbf{H}}_{\mathrm{LS}} \in \mathbb{R}^{\sqrt{M} \times \sqrt{M}\times  2} $,  which is fed into a 2D convolutional layer with $F$ filters and a kernel size of $3 \times 3$. This convolutional layer serves for shallow feature extraction and dimensionality enhancement ${\mathbf{H}}_{{1}} \in \mathbb{R}^{\sqrt{M} \times \sqrt{M}\times  F} $. The resulting feature maps are subsequently reshaped into a matrix of size ${\mathbf{H}}_{{2}} \in \mathbb{R}^{M\times  F} $, which is used as the input to the Dual-Attention Embedding module.





\subsection{Dual-Attention Embedding}\label{AA}
The Dual-Attention Embedding module is specifically designed for preliminary feature extraction prior to the LLM. As illustrated in Fig.~\ref{figure2}, it consists of two Parallel Feature-Spatial Attention branches, a fully connected (FC) layer, and a positional encoding module.

\begin{figure}[htbp]
	\centering
	\includegraphics[width=0.47\textwidth]{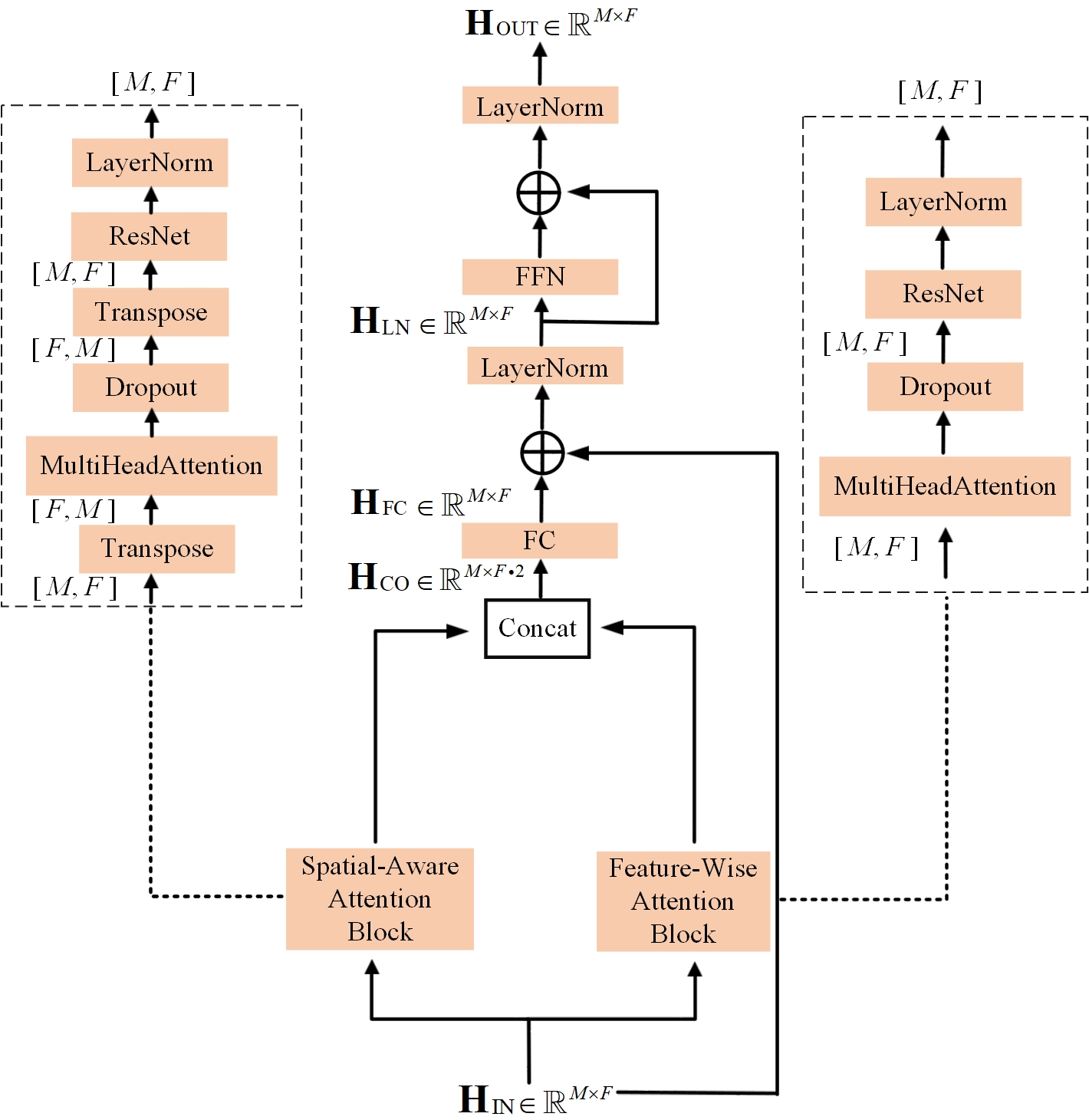}
	\caption{Illustration of Parallel Feature-Spatial  Attention module.}
	\label{figure3}
\end{figure}

 As illustrated in Fig.~\ref{figure3}, the input feature $\mathbf{H}_{\mathrm{IN}}$ is first processed in parallel by a Feature-Wise Attention module and a Spatial-Aware Attention module. The outputs of these two branches are concatenated along the last dimension to form $\mathbf{H}_{\text{CO}}$, effectively integrating spatial and feature representations. This fusion facilitates the ability of the subsequent LLM to learn the spatial configuration of antenna elements.
The fused representation $\mathbf{H}_{\text{CO}}$ is then passed through a fully connected layer to restore the original feature dimensionality, denoted as $\mathbf{H}_{\text{FC}}$. A residual connection with the initial input $\mathbf{H}_{\text{IN}}$ is applied, followed by layer normalization to obtain $\mathbf{H}_{\text{LN}}$, which helps mitigate gradient vanishing or explosion and improves model stability.
The output $\mathbf{H}_{\text{LN}}$ is further refined by a feed-forward network (FFN), along with another residual connection and layer normalization, yielding the final output $\mathbf{H}_{\text{OUT}}$.

The representation obtained after the second Parallel Feature-Spatial Attention layer is denoted as ${\mathbf{H}}_{{3}} \in \mathbb{R}^{M\times F} $. The communication feature matrix $\mathbf{H}_3$ passes through a fully connected layer to project its last dimension to $d$, the embedding size required by the GPT2\cite{b15} model. The resulting tensor, denoted as ${\mathbf{H}}_{{4}} \in \mathbb{R}^{M\times  d} $, is enhanced with a learnable positional embedding following the GPT-2 design, resulting in ${\mathbf{H}}_{{5}} \in \mathbb{R}^{M\times  d} $, which serves as the input to the LLM.

The Feature-Wise Attention module employs a multi-head attention mechanism, where multiple heads capture diverse feature representations that are subsequently fused into a unified output.
For each head $i= \{1,\cdots,I \}$, we define query matrices $\mathbf{Q}_{ i} = \mathbf{H}_{\text{IN}}\mathbf{ W}_{ i}^\mathrm{Q}$ , key matrices $\mathbf{K}_{ i} = \mathbf{H}_{\text{IN}}\mathbf{ W}_{ i}^\mathrm{K}$, and value matrices $ \mathbf{V}_{ i} = \mathbf{H}_{\text{IN}}\mathbf{ W}_{ i}^\mathrm{V}$ , where $\mathbf{Q}_i$, $\mathbf{K}_i$, and $\mathbf{V}_i \in \mathbb{R}^{M \times d_k}$. Specifically, $d_k =\frac{F}I$ represents the hidden dimension of each head, however, in this work, we intentionally set $d_k = F$ to facilitate deeper feature extraction from long sequences. The output of the multi-head attention mechanism is obtained by computing the attention for each head as
\begin{equation}
	\mathbf{Z}_i = \text{Attention}(\mathbf{Q}_i,\mathbf{K}_i,\mathbf{V}_i),	
\end{equation}
by aggregating each $\mathbf{Z}_i \in \mathbb{R}^{M \times d_k}$ in every head, the outputs of all heads are then concatenated, and a linear transformation is applied to obtain the final result
			\begin{equation}
	\mathbf{ Z} =  \text{Concat}(\mathbf{ Z}_1, \mathbf{ Z}_2, \dots, \mathbf{ Z}_I) \mathbf{W}^\mathrm{O},	
\end{equation}
	where $\mathbf{ Z}\in \mathbb{R}^{M \times F}$, $\text{Concat}(\mathbf{ Z}_1, \mathbf{ Z}_2, \dots, \mathbf{ Z}_I )\in \mathbb{R}^{M \times I\cdot d_k}$ and $\mathbf{W}^\mathrm{O}\in \mathbb{R}^{I\cdot d_k \times F}$. The out $ \mathbf{ Z}$
	represents the aggregated representation that integrates the feature dimensions from all attention heads, which is used for subsequent fusion with Spatial-Aware Attention.

For the Spatial-Aware Attention module, the input is transposed to align spatial dimensions, followed by multi-head attention to extract spatial features. The outputs are then fused into a unified spatial representation.

Unlike conventional methods that rely solely on feature attention, this work introduces a Parallel Feature-Spatial Attention module, motivated by the structured physical arrangement of antennas in a uniform linear array. While feature attention is effective at modeling semantic representations of the channel, it is limited in capturing the spatial dependencies among antenna elements. In contrast, spatial attention explicitly models the spatial correlations inherent in the array structure, thereby providing the LLM with physical priors. By integrating spatial attention with feature attention, the model's awareness of array geometry is enhanced, leading to improved channel estimation accuracy.

To incorporate positional information of the input signals, this work adopts the learnable positional encoding mechanism from GPT-2. Specifically, for each input patch vector $\mathbf{H}_j$, a trainable positional encoding $\mathbf{P}_j$ corresponding to its position is added, resulting in the final input representation
\begin{equation}
\tilde{\mathbf{H}}_j = \mathbf{H}_j + \mathbf{P}_j,
\end{equation}
where, $\mathbf{H}_j \in \mathbb{R}^d$ denotes the embedding of the $j$-th input token $\mathbf{H}_{4}$, and $\mathbf{P}_j \in \mathbb{R}^d$ is the learnable positional encoding vector corresponding to position $j$. The dimension $d$ the feature dimension of the GPT-2 model, corresponding to its hidden size.

\subsection{Partially Training LLM}
Recent studies\cite{b9},\cite{b10} show that LLMs, pretrained on large text corpora, can be fine-tuned for cross-modal tasks like channel prediction and beam selection. Inspired by their strong generalization ability, we explore applying LLMs to channel estimation.
 However, due to the significant difference between text data and the channel matrix, pretrained LLMs cannot directly process non-linguistic data. To address this, we design a preprocessor and embedding module to convert the channel matrix into an embedded representation. The resulting embedded ``tokens" $\mathbf{H}_5$ are then fed into the backbone of the LLM for modeling,
\begin{equation}
	\mathbf{H}^{\mathrm{LLM}} = \mathrm{LLM}(\mathbf{H}_{{5}}) \in \mathbb{R}^{M \times d},
\end{equation}
where LLM(·) denotes backbone networks of the LLM. Without loss of generality, this study adopts GPT-2 as the backbone architecture of the LLM. The backbone of GPT-2 consists of a learnable positional embedding layer and a stack of Transformer decoder blocks, where the number of layers and the feature dimensions can be flexibly adjusted based on specific requirements.

During training, layers 1 through 10 of GPT-2 are fully frozen, including the multi-head attention mechanisms, feed-forward networks, and layer normalization layers, in order to retain general pre-trained knowledge. Fine-tuning is restricted to the multi-head attention, feed-forward, and layer normalization components in layers 11 and 12, as well as the final output layer normalization and the positional embedding, to better adapt the model to the channel estimation task.

It is worth noting that the GPT-2 backbone in this framework can be readily replaced with other LLMs. The selection of the LLM type and model size should balance training cost and performance requirements.
\subsection{Postprocessing \& Output}
The output of the LLM, denoted as ${\mathbf{H}}_{{6}} \in \mathbb{R}^{M\times  d} $, is first passed through a fully connected layer to reduce the last dimension to $F$,
\begin{equation}
\mathbf{H}_7 = \text{FC}({\mathbf{H}_6}).
\end{equation}
Where, $\mathbf{H}_7 \in \mathbb{R}^{ M \times F}$, is processed sequentially by two 2D convolutional layers with 64 filters each, followed by a third 2D convolutional layer with 2 filters, progressively refining and stabilizing the feature representation. The final output $\mathbf{ {H}}_8 \in \mathbb{R}^{\sqrt{M} \times \sqrt{M}\times  2} $ is interpreted as the estimated noise component in the channel.
To obtain the final estimated channel  $\hat{\mathbf{ {H}}} \in \mathbb{R}^{\sqrt{M} \times \sqrt{M}\times  2} $, the predicted noise ${\mathbf{H}}_{{8}}  $ is subtracted from the noisy input channel $\hat{\mathbf{H}}_{\mathrm{LS}}$,
\begin{equation}
\hat{\mathbf{H}} = \hat{\mathbf{H}}_{\mathrm{LS}} - \mathbf{H}_8. 
\end{equation}

The final estimated channel vector $\hat{\mathbf{ {h}}} \in \mathbb{C}^{ {M}} $
is obtained by reshaping the estimated matrix $\hat {\mathbf{H}}$.
This subtraction-based residual modeling helps suppress noise and recover a cleaner channel representation.
\subsection{Training Settings}
 In XL-MIMO systems, user channels may shift between near-field and far-field due to mobility. To avoid performance loss from model mismatch, a unified channel estimation framework is needed that adapts to varying propagation conditions without requiring dynamic state information. To this end, LLM4XCE is trained on data from a hybrid-field channel model\cite{b16}. According to equation (6), the channel vector $ \mathbf{h}$ is power-normalized to enhance the stability and convergence of neural network training. Selecting appropriate values for $L_0$
and $L$ further improves the training efficiency and generalization ability of LLM4XCE under varying propagation conditions and path counts. Given $L_0$ and $L$, the training dataset $\mathcal{D}_{\text{tr}}$ is generated using the channel vector $\mathbf{h}$ as defined in equation (6), and LLM4XCE is trained on this dataset by minimizing the mean squared error (MSE) loss,
\begin{equation}
\mathcal{L} = \frac{1}{N_{\mathrm{tr}}} \sum_{j=1}^{N_{\mathrm{tr}}} \left\| {{\mathbf{H}}}^{(j)} - \hat{\mathbf{H}}^{(j)} \right\|_F^2,
\end{equation}
where, $N_{\text{tr}}$ denotes the number of training samples, and the superscript ($j$) indicates the $j$-th sample.
\section{Experiments}
In this section, we first present the preparations and data processing conducted for the simulation. The second part provides the simulation results, comparing and analyzing LLM4XCE with other approaches.
\subsection{Simulation Setup}



The simulation parameters are configured as follows the number of BS antennas $ M=256$, the wavelength $ \lambda=0.01m$, the mean path gain $ \sigma^2=1$, $\theta_l \sim U\left( -\frac{\pi}{2}, \frac{\pi}{2} \right) $, and $ r_l \sim U(10, 80)$ meters. The normalized MSE (NMSE), $\mathbb{E}\{ \|{{\mathbf{H}}} - \hat{{\mathbf{H}}}\|_F^2 / \|{{\mathbf{H}}}\|_F^2 \}
$, is used to measure the channel estimation performance.
   In the simulation, the training and validation datasets, generated based on the hybrid-field channel model in (6) with $L = 6$ and $L_0 = 1$, consist of 45,000 and 5,000 samples, respectively. The test set, also based on the hybrid-field channel model in (6), consists of 2,000 samples.
 \begin{table}[htbp]
	\caption{Hyper-parameters for network training}
	\begin{center}
		\begin{tabular}{cc}  
			\hline  
			\textbf{Parameter} & \textbf{Value}  \\
			\hline  
			Batch size  &64   \\
			
			Epochs  & 200  \\
			Optimizer &  Adam (betas = (0.9,0.999)) \\
			Starting learning rate   & 0.001  \\
			Adaptive LR decay    & 0.1 per 50 epochs      \\
			\hline  
		\end{tabular}
	\end{center}
	\label{tab1}
\end{table}
\begin{figure}[t]
	\centering
	\includegraphics[width=9cm, height=6cm]{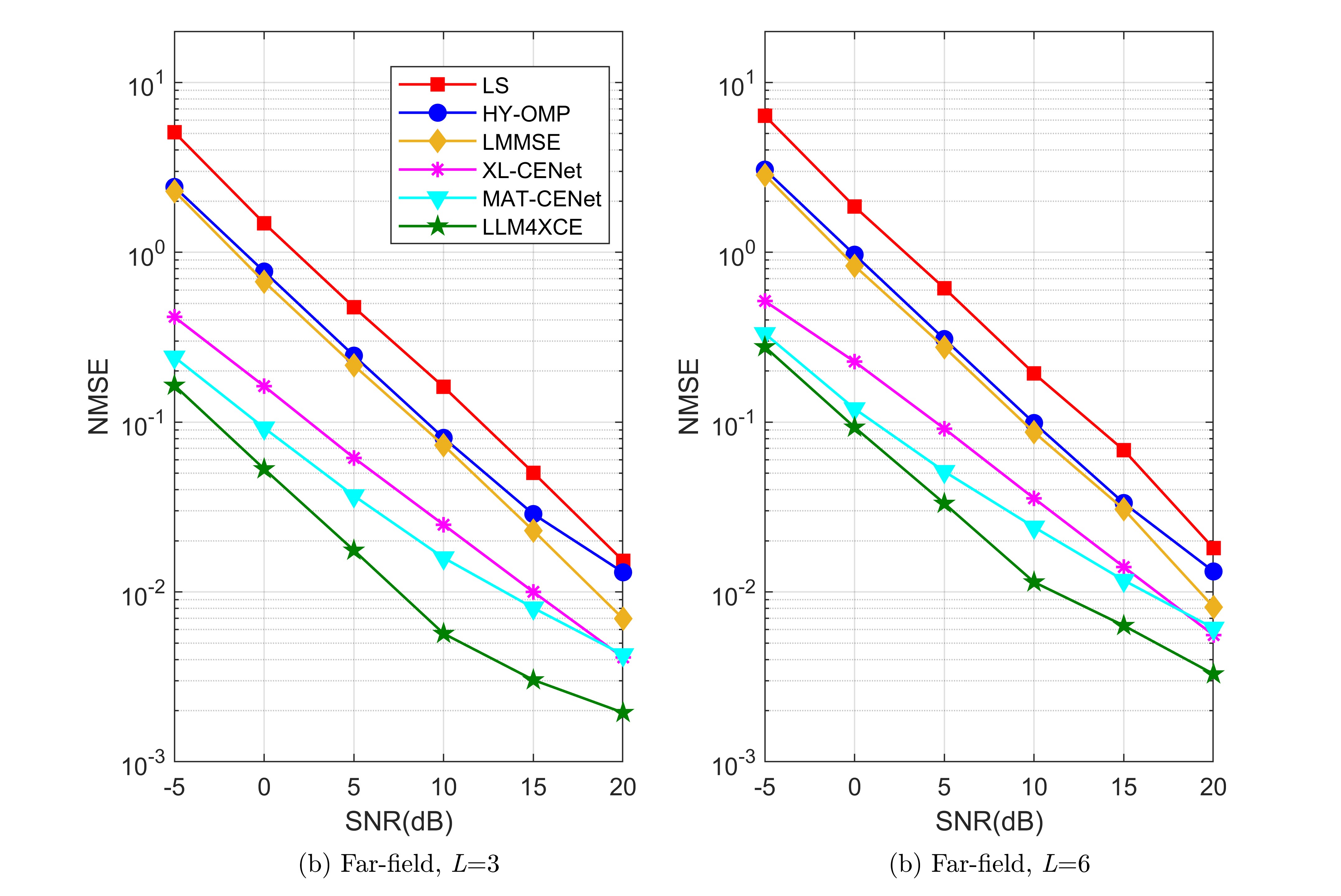}  
	\caption{NMSE versus SNR for the far-field user.}
	\label{figure4}
\end{figure}
In this work, the smallest version of GPT-2 with a feature dimension of $d = 768$ is employed. For downstream processing, only the top two Transformer layers are activated. The LLM4XCE network comprises 17M trainable parameters, with an additional 109M parameters kept frozen during training. The hyperparameters used for model training are summarized in Table~\ref{tab1}.
 Finally, we compare the proposed method with traditional LS estimation, linear minimum mean square error (LMMSE) estimation, hybrid field orthogonal matching pursuit (HY-OMP)\cite{b16}, XL-MIMO channel network (XLCENet)\cite{b4}, and mixed attention transformer based channel estimation neural network (MAT-CENet)\cite{b6}.
\subsection{Simulation Results}
The following discusses the channel estimation performance of LLM4XCE versus other schemes under various signal-to-noise ratio (SNR) conditions. The horizontal axis shows SNR values from -5 dB to 20 dB, “LLM4XCE”, “MAT-CENet”, and “XLCNet” are trained on the same dataset with 1 far-field and 5 near-field paths.

Fig.~\ref{figure4} illustrates the normalized mean squared error (NMSE) performance of far-field user under different SNR values, corresponding to scenarios with 3 and 6 far-field paths. As shown in Fig.~\ref{figure4}, LLM4XCE demonstrates significant advantages in far-field environments, consistently outperforming all baseline methods. Notably, it achieves higher sensitivity and accuracy when the number of paths is 3. Even in more complex scenarios with 6 paths, the model maintains stable and superior estimation performance.

\begin{figure}[t]
	\centering
	\includegraphics[width=9cm, height=6cm]{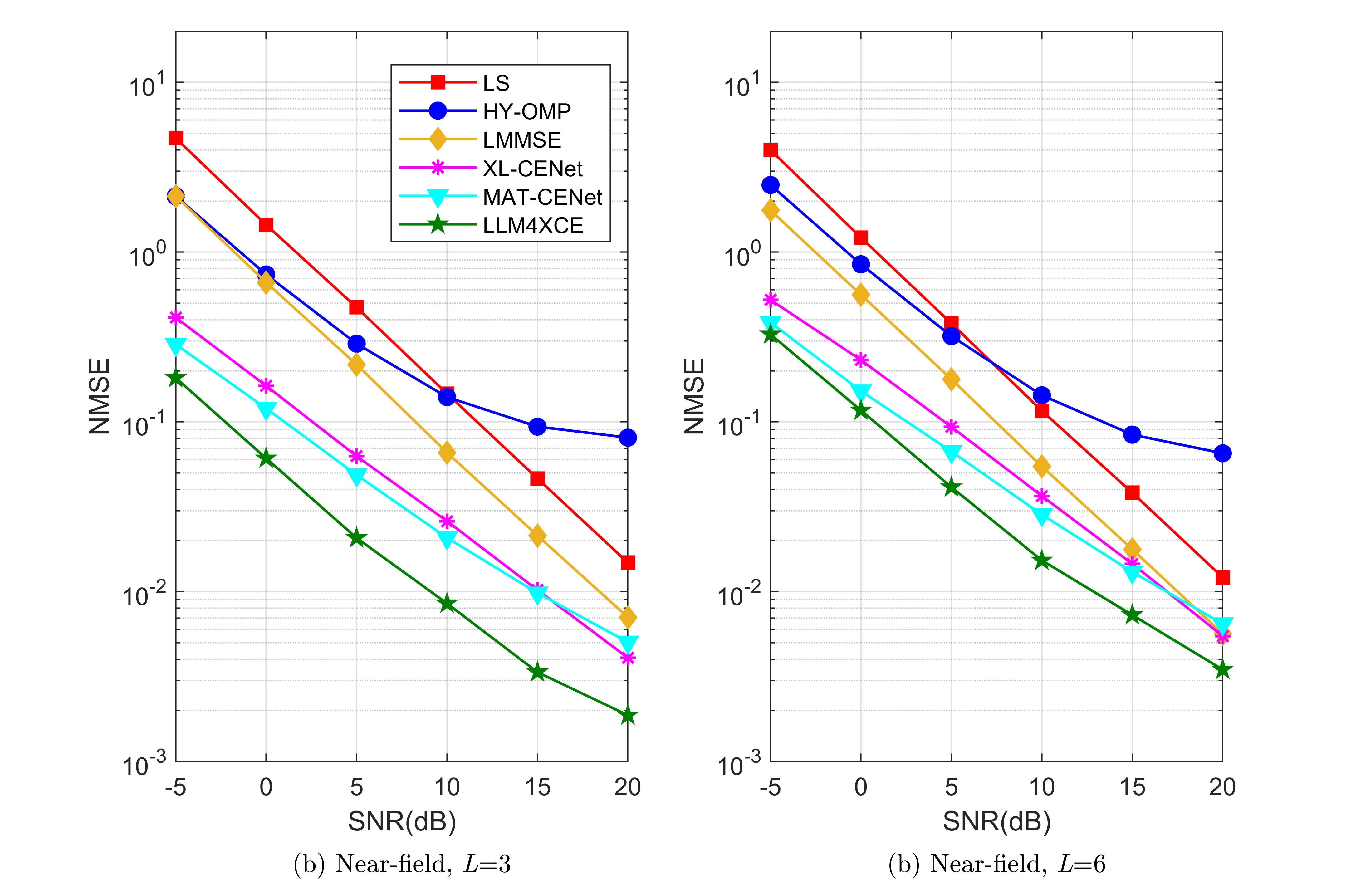}  
	\caption{NMSE versus SNR for the near-field user.}
	\label{figure5}
\end{figure}
\begin{figure}[t]
	\centering
	\includegraphics[width=8cm, height=6cm]{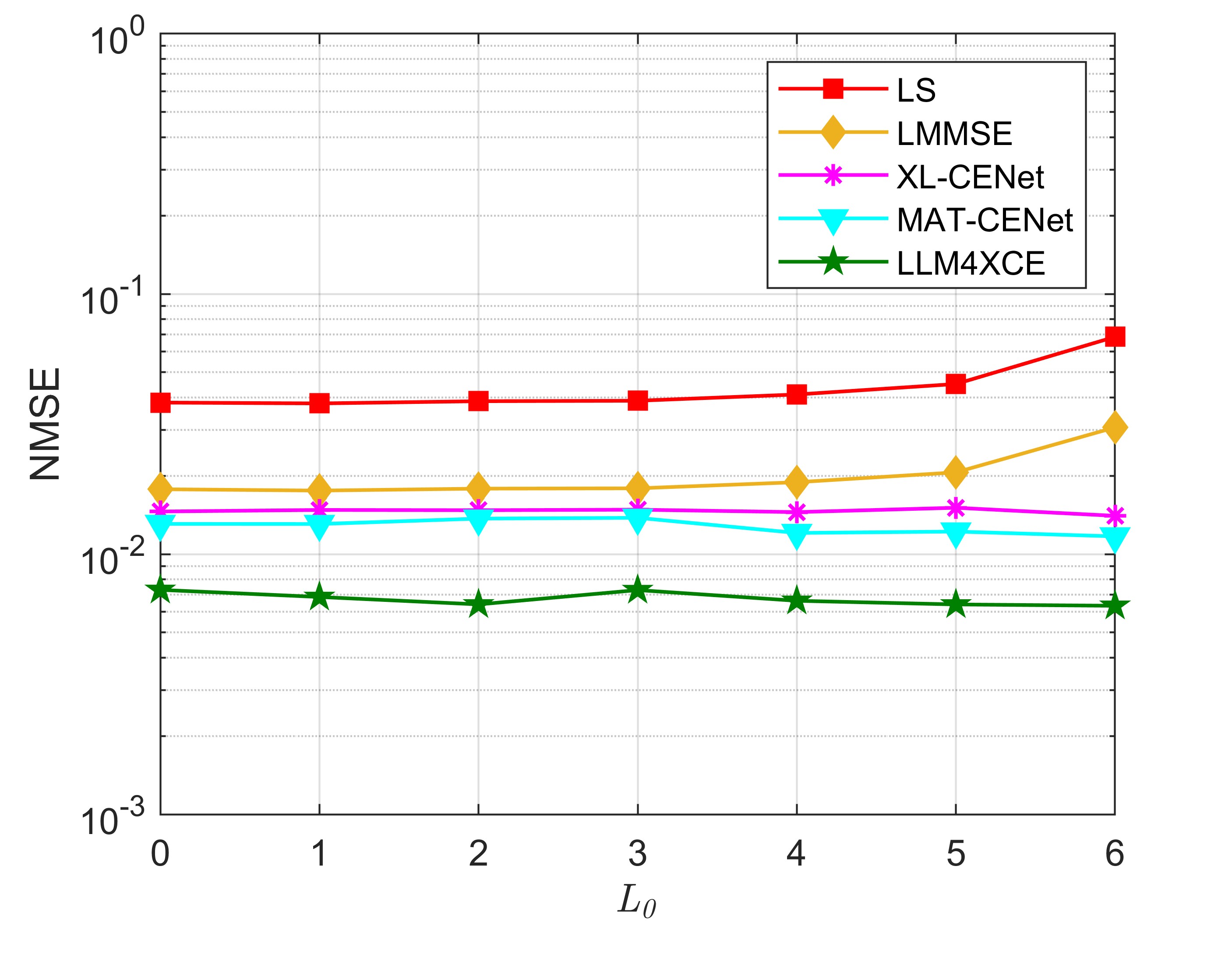}  
	\caption{NMSE versus the number of far-field paths $L_0$ for the hybrid-field
		with $L = 6$.}
	\label{figure6}
\end{figure}

Fig.~\ref{figure5} illustrates the NMSE performance trends of near-field user with varying SNR values under two different path settings: 3 and 6 near-field paths. The results show that LLM4XCE consistently outperforms existing baseline methods across all SNR levels, demonstrating its strong capability in extracting features under complex near-field channel conditions. Although the increase in the number of paths leads to a slight rise in estimation error, LLM4XCE still maintains a significant performance advantage in high-SNR scenarios, further highlighting its robustness in high-quality communication environments.

Fig.~\ref{figure6} compares the NMSE performance of different methods in a hybrid-field scenario with a fixed SNR of 15dB and a total of 6 propagation paths, as the number of far-field paths increases. As shown in Fig.~\ref{figure6}, LLM4XCE consistently outperforms all baseline methods regardless of the far-field path ratio, demonstrating strong robustness and adaptability in hybrid-field environments. Compared with MAT-CENet, the estimation error is reduced by a factor of 2. Notably, since the training data were generated using a configuration of 1 far-field path and 5 near-field paths, the model achieves optimal performance under this setting, indicating its effective learning of the training distribution. Moreover, thanks to the powerful representation learning of LLMs, LLM4XCE maintains high estimation accuracy even when test conditions differ from training. This demonstrates strong generalization
and effectiveness in modeling complex channels for large-scale
MIMO estimation tasks.
			
			
			

\section{Conclusion}
In this article, we proposed LLM4XCE, a novel channel estimation framework for hybrid-field XL-MIMO systems leveraging LLMs. A Parallel Feature-Spatial Attention module fuses pilot features from spatial and semantic views, adapting them for LLM input. To reduce computation and enhance performance, only the top two Transformer layers are fine-tuned. Simulation results show that our method outperforms existing advanced approaches in estimation accuracy and generalization ability, fully demonstrating the effectiveness of LLM-based solutions in enhancing channel estimation performance and integrating semantic communication advantages in complex wireless environments.
\emph{}

\end{document}